\documentclass{Interspeech}
\usepackage{float}
\usepackage{multirow}

\interspeechcameraready

\title{Leveraging LLM and Self-Supervised Training Models for Speech Recognition
in Chinese Dialects: A Comparative Analysis}

\author[affiliation={1,2}]{Tianyi}{Xu}
\author[affiliation={1}]{Hongjie}{Chen}
\author[affiliation={1}]{Qing}{Wang}
\author[affiliation={1}]{Hang}{Lv}
\author[affiliation={1}]{Jian}{Kang}
\author[affiliation={1}]{Jie}{Li}
\author[affiliation={2}]{Zhennan}{Lin}
\author[affiliation={1}]{Yongxiang}{Li} 
\author[affiliation={2}]{Lei}{Xie}

\affiliation{Institute of Artificial Intelligence (TeleAI)}{China Telecom}{China}
\affiliation{Audio, Speech and Language Processing Group (ASLP@NPU), School of Computer Science}{Northwestern Polytechnical University}{China}
    \email{xutianyi@mail.nwpu.edu.cn; lxie@nwpu.edu.cn}

\usepackage{comment}

\begin{document}

\maketitle

\begin{abstract}
    
Large-scale training corpora have significantly improved the performance of ASR models. Unfortunately, due to the relative scarcity of data, Chinese accents and dialects remain a challenge for most ASR models. Recent advancements in self-supervised learning have shown that self-supervised pretraining, combined with large language models (LLM), can effectively enhance ASR performance in low-resource scenarios. We aim to investigate the effectiveness of this paradigm for Chinese dialects. Specifically, we pre-train a Data2vec2 model on 300,000 hours of unlabeled dialect and accented speech data and do alignment training on a supervised dataset of 40,000 hours. Then, we systematically examine the impact of various projectors and LLMs on Mandarin, dialect, and accented speech recognition performance under this paradigm. Our method achieved SOTA results on multiple dialect datasets, including Kespeech. We will open-source our work to promote reproducible research.
\footnote{Work done during internship at China Telecom.}


    
\end{abstract}

\section{Introduction}
Large-scale training corpora have significantly improved the performance of automatic speech recognition (ASR) models~\cite{whisper,zhang2023google,pratap2024scaling, chen2022wavlm}. However, labeled dialect data remains prohibitively expensive, making it challenging to obtain in large quantities. As a result, most pre-trained models in Chinese speech processing communities, such as WeNet~\cite{wenet}, K2~\cite{yao2023zipformer}, and FunASR~\cite{gao23g_interspeech}, exhibit limited capabilities in dialect recognition. Fortunately, advances in self-supervised learning have demonstrated the potential of large-scale training to further improve model performance. Self-supervised large language models (LLM) ~\cite{floridi2020gpt, touvron2023llama,  achiam2023gpt, team2024gemini}have shown exceptional capabilities to understand and generate human language, and self-supervised speech models such as Wav2Vec2~\cite{baevski2020wav2vec}, HuBERT~\cite{hubert} and Data2Vec2~\cite{baevski2022data2vec} have demonstrated strong proficiency in modeling speech. Given that ASR traditionally relies on both acoustic and language modeling, the advantages of self-supervised pre-trained acoustic models and LLMs—particularly their scalability in terms of training data and model size—present promising opportunities for low-resource ASR systems.

Research efforts to integrate LLMs with ASR systems generally follow two primary approaches. The first is the cascaded approach~\cite{huang2024audiogpt, dighe2024leveraging, ma2023can}, where LLMs are employed to refine the n-best hypotheses generated by ASR systems. Among studies adopting this approach, MMGER~\cite{mu2024mmger} stands out as a closely related work, achieving the previous SOTA performance in Chinese accents on the Kespeech~\cite{tang2021kespeech} dataset. However, this approach has a notable limitation: for complex cases, the n-best hypotheses may fail to provide sufficiently informative input to the LLM, making it difficult for the model to infer correct results from erroneous predictions. This limitation can sometimes result in semantic distortions during error correction.
The second approach involves training audio-text cross-modal LLMs~\cite{salmonn,simple-asr-llm,qwen-audio}, where a speech encoder processes audio signals to generate embeddings that are subsequently fed into an LLM for decoding. This method aims to integrate acoustic features more effectively with linguistic context to enhance ASR accuracy. Specifically, SALMONN~\cite{salmonn} employs Whisper~\cite{whisper} to extract semantic content while utilizing BEATs~\cite{pmlr-v202-chen23ag} to enable comprehensive perception of speech, music, and sound events. Qwen-Audio and Qwen2-Audio~\cite{qwen-audio,chu2024qwen2} use Whisper as an encoder and optimize task performance across various domains with structured task instructions. Similarly, SLAM-ASR~\cite{simple-asr-llm} achieved SOTA results on the 960-hour English LibriSpeech dataset by training only the projector. In the context of Chinese ASR, Geng et al.~\cite{geng2024unveiling} achieved SOTA performance on AISHELL, Test-net, and Test-meeting by leveraging HuBERT encoder and Transformer projection layers.
\begin{figure*}[]
\vspace{-1cm}
\centering
\resizebox{0.9\textwidth}{!}{\includegraphics{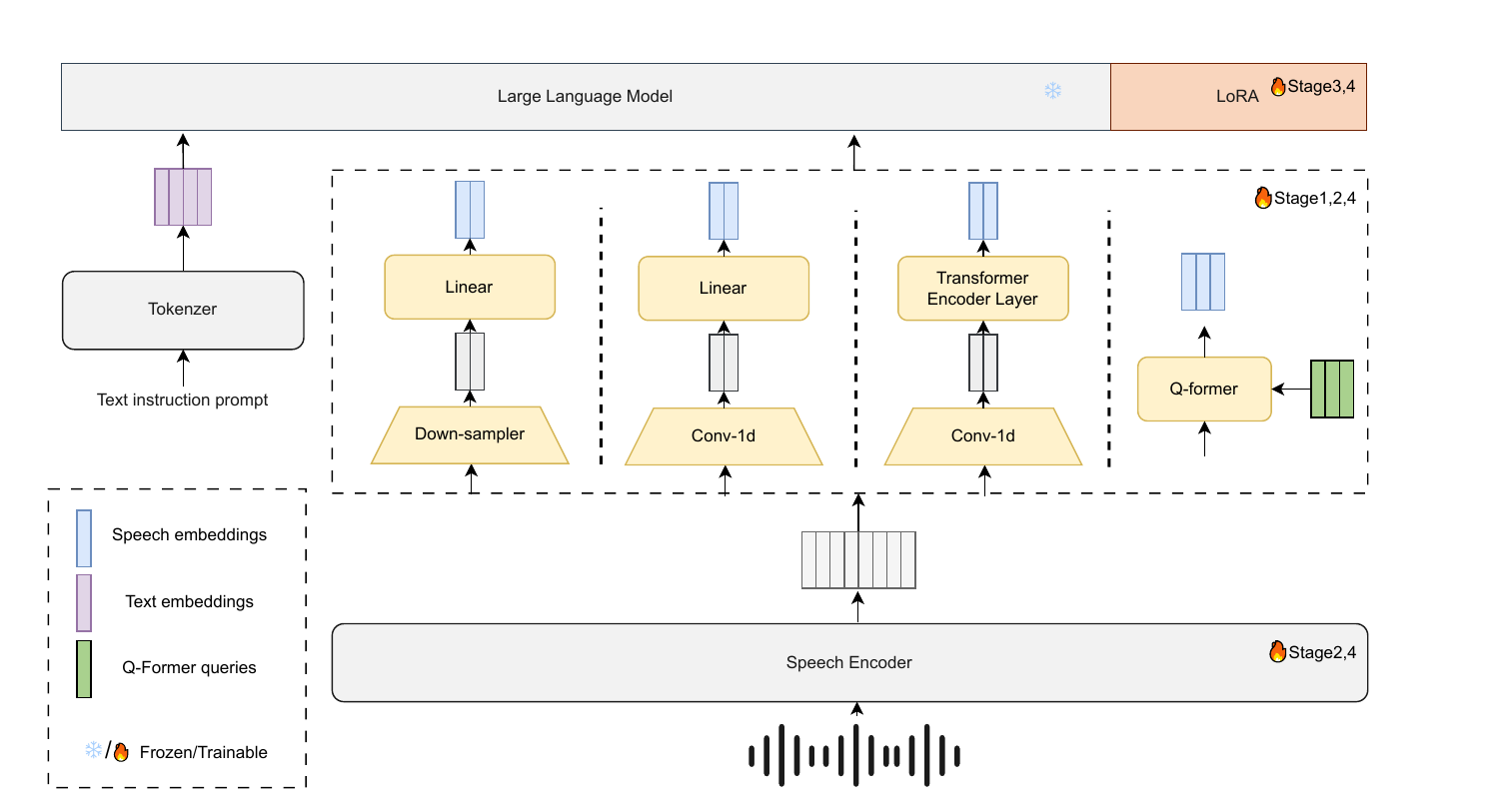}}
\caption{Overall model structure.}
\vspace{-0.5cm}
\label{fig:0001}
\end{figure*}

Our study extends these investigations by exploring the potential of the speech foundation encoder with the LLM-Decoder paradigm on large-scale Chinese dialect datasets. Specifically, we utilize two datasets: an unlabeled dataset consisting of approximately 300,000 hours of dialect and accented speech and a labeled dataset comprising 40,000 hours. Using these resources, we systematically investigate the impact of various speech encoders, projectors, and LLMs on the ASR performance of Chinese dialects.
Our experimental results reveal several important findings:
\begin{itemize}
    \item First, among speech encoders, Data2Vec2 is a good speech encoder for large language models and can have competitive performance when downsampled to 25Hz. Combined with the 4-time downsampling from the projector, it can achieve competitive performance at 6.25 Hz framerate for the LLM, greatly reducing the computational overhead.   
    \item Second, fully connected layers outperform other projectors, particularly during the early stages of training. Notably, differences in the effectiveness of projectors are most prominent during the first training stage but diminish as the parameters of the pre-trained models are progressively unfrozen in subsequent stages.  
    \item Third, a multi-stage training strategy proves highly effective in enhancing model performance, with the addition of a fourth training stage, building upon conventional three-stage approaches, yields                 
                further improvements. 
    
\end{itemize}

Compared to previous work~\cite{simple-asr-llm,geng2024unveiling}, our approach employs Data2Vec2 as the speech encoder and incorporates convolutional layers as the projector. Combined with a four-stage training strategy, the proposed method achieves SOTA performance on multiple Chinese dialect and accent test sets, including Kespeech (with CER of 6.48\%
vs. 7.52\%), even with a small LLM (0.5B) and a low frame rate (6.25Hz).
To promote reproducibility, we will release all associated training, inference, and evaluation scripts, as well as pre-trained models and training logs.

\section{Method}

\label{sec:2.1}
As illustrated in Figure~\ref{fig:0001}, the model architecture consists of an audio encoder and a LLM. For each sample during training, we define the text prompt (e.g., "Transcribe the following speech"), the speech utterance, and the corresponding transcription as \( P \), \( S \) and \( T \), respectively.  
The text prompt and transcription are tokenized using the LLM's tokenizer and embedded into feature vectors  \( E_p \) and \( E_t \) as:
\begin{align}
E_p &= \text{Embedding}(\text{Tokenizer}(P)),\label{eq:multi1} \\
E_t &= \text{Embedding}(\text{Tokenizer}(T)).\label{eq:multi2}
\end{align}
For the input speech \( S \), features are first extracted using the speech encoder, producing the encoder output \( H_s \), represented as:
\begin{equation}
H_s = \text{Encoder}(S).
\label{myequation1}
\end{equation}
The encoder output \( H_s \) is then passed through a projector, followed by a linear layer, to generate a feature sequence \( E_{\text{s}} \) with the same dimensionality as the input to the LLM, denoted as:
\begin{equation}
E_{\text{s}} = \text{Linear}(\text{Projector}(H_s)),
\label{myequation2}
\end{equation}
Here, the projector preserves the dimensionality of the features output by the speech encoder, while the linear layer maps these features to the embedding space of the LLM.
Finally,\( E_{\text{s}} \),  \( E_p \), and \( E_t \) are concatenated into a unified feature sequence and fed into the LLM. The model generates the transcription \(Y\), as the output, defined as:
\begin{equation}
Y = \text{LLM}(\text{Regulation}(E_p, E_{\text{s}}, E_t)).
\label{eq:1}
\end{equation}



\section{Experimental Setup}
\subsection{Datasets}
\textbf{Training Set:} The training data comprises a diverse set of internal dialect and accent datasets, including samples from Anhui, Gansu, Hebei, Shandong, Shanxi, Tianjin, Cantonese, Henan, Sichuan, Chongqing, Northeast China, Shaanxi, Hubei, Fujian, Guizhou, Hangzhou, Hunan, Jiangxi, Shanghai, Suzhou, Yunnan, Chongqing and Hakka. Additionally, public Mandarin datasets, such as Wenetspeech~\cite{wenetspeech}, Aishell~\cite{aishell1}, and AliMeeting, are incorporated. For training the unsupervised models, the total dataset size is approximately 300,000 hours, while the supervised dataset is 40,000 hours.

\noindent \textbf{Testing Set:} Due to the lack of open-source test datasets for many dialects and the limitations of existing datasets—such as overly short utterances that fail to capture distinct dialectal characteristics—we employ internally annotated test sets for evaluation. Specifically, we conduct evaluations on four dialects: Shanghai, Hunan, Henan, and Cantonese, each with approximately 2,000 audio samples. Additionally, we utilize the open-source Kespeech dataset~\cite{tang2021kespeech} to assess performance on accented speech. Kespeech includes Mandarin as well as dialects from regions such as Beijing, Jiang-Huai, Jiao-Liao, Ji-Lu, Lan-Yin, Northeastern, Southwestern, and Zhongyuan. However, we observe that the accents in this dataset are not particularly pronounced, likely because regional labels for speakers were directly assigned as accent labels, even when many speakers exhibit minimal accent features. Finally, to evaluate performance on 
Mandarin, we use the Test-net and Test-meeting subsets from the Wenetspeech dataset.


\subsection{Components}
\subsubsection{LLM}
To ensure robust recognition performance across various Chinese accents and dialects, we require an LLM with good performance in Chinese-related tasks. Given the complexity of LLMs, the academic community currently lacks standardized evaluation metrics. Therefore, we refer to the CLiB~\footnote{https://github.com/jeinlee1991/chinese-LLM-benchmark?tab=readme-ov-file} leaderboard, which assesses the capabilities of Chinese LLMs across multiple tasks. Among open-source models of under 7B parameters, the Qwen 2 series~\cite{yang2024qwen2} achieved the highest scores. Consequently, our study utilizes Qwen 2 of 0.5B, 1.5B, 7B, and Qwen 2.5 3B (as Qwen 2.5 was only released during the later stages of our research). Notably, our testing reveals that the Qwen series also demonstrates the capability to handle conversational responses in dialectal and colloquial Chinese expressions.


\subsubsection{Speech foundation encoder}
In prior work~\cite{chen2024telespeechpt},  the performance of several self-supervised speech encoders was investigated, including Wav2vec2~\cite{baevski2020wav2vec}, Hubert~\cite{hubert}, BEST-RQ~\cite{chiu2022self}, DinoSR~\cite{liu2023dinosr}, and Data2vec2~\cite{baevski2022data2vec}. Our findings showed that, in Wenetspeech pre-training, Data2vec2 outperformed the other models. Subsequently, we trained Data2vec2 on a larger dataset consisting of 300,000 hours of unlabeled multilingual and accented speech. For the current study, we use the TeleSpeechPT variant of Data2vec2 with a model size of 24 layers and 301 million parameters. The model was trained at a sampling rate of 25Hz, which is lower than the commonly used 50Hz rate of the Whisper encoder.


\subsubsection{Projector}
The speech modality is more sparsely represented compared to the text modality, leading to increased computational complexity. To enhance inference efficiency, previous studies have frequently employed projection layers to align text and speech modalities. Since the community lacks a consensus conclusion on the most suitable projection layer for speech-based LLMs, we perform a comparative evaluation in the context of Chinese dialects. Specifically, we compare four different projection layers: fully connected (Linear), 1-dimension Convolutional (Conv1d), Transformer, and Q-Former~\cite{yu2024connecting}. For the Linear and Conv1d layers, we follow the configuration from SLAM and conduct experiments with downsampling rates of 1, 2, 4, and 8. The Q-Former configuration also follows SLAM, with the number of queries set to 64. The Transformer configuration follows the setup in~\cite{geng2024unveiling}.

\subsection{Training strategy}
\label{sec:3.1}
LLM-based models are typically very large, and training the entire model simultaneously can sometimes lead to suboptimal performance.  Previous works~\cite{salmonn,qwen-audio,geng2024unveiling} used a three-stage fine-tuning strategy to ensure convergence between different modules. However, we hypothesize that this fine-tuning approach might cause the model to converge to local optima within individual modules rather than achieving a global optimum. Therefore, building on the three-stage training framework, we introduce an additional fourth stage, which involves fine-tuning the entire model. Our experiments shows that this modification improve ASR performance. Specifically, the training process is shown in Figure~\ref{fig:0001}: (1) First, only the projector is trained while all other components are frozen. (2) Second, only the speech encoder is trained. (3) Third, the encoder and projector are frozen, and LoRA is used to fine-tune the LLM. (4) Finally, both the encoder and the projector are unfrozen for further optimization.

\begin{table}[]
    \centering
    \caption{Comparison of CER (\%) on Test-net/Test-meeting for Different Projection Layers after Pre-trained only Data2vec2 encoder vs CTC fine-tuned Data2vec2 encoder and Qwen2 0.5B LLM during the First Training Stage.}
    \scalebox{0.85}{
\begin{tabular}{cccc}
\hline
            & Frame Rate & Pretrained          & Finetuned            \\ \hline
Conv1d      & 6.25Hz     & 18.19/16.64         & 31.46/27.87          \\
Linear      & 6.25Hz     & \textbf{16.47/8.80} & \textbf{14.09/11.09} \\
Transformer & 6.25Hz     & 17.32/12.37         & 14.34/11.32          \\
Q-former    & -          & 27.00/17.76            & 23.72/17.53          \\ \hline
\end{tabular}
  
    \label{tab:projectortest}
    }
\vspace{-0.3cm}
\end{table}

\begin{table}[]
  \centering
    \caption{Comparison of CER (\%) on Test-meeting/Test-net after fine-tuning on Wenetspeech with Multiple Projection Layers at Different Sampling Rates.}
    \scalebox{0.85}{
    \begin{tabular}{ccccc}
    \hline
                & 25Hz                         & 12.5Hz      & 6.25Hz      & 3.125Hz     \\ \hline
    Conv1D      & \multirow{2}{*}{\textbf{16.48/10.79}} & 20.44/14.10 & 31.46/27.87 & 38.97/41.66 \\
    Linear      &                              & \textbf{18.15/14.89} & \textbf{14.09/11.09} & 
\textbf{24.62/25.00} \\
    Transformer & 17.32/12.37                  & 20.03/14.57 & 19.06/11.87 & 27.65/26.94 \\ \hline
  \end{tabular}
    }
    \label{tab:projectorsrtest}
\vspace{-0.5cm}
\end{table}
\subsection{Implementation details}

We train our model using 8 40G A100 GPUs, and fine-tuning for the 7B version takes about 30 days for four stages. We use the framework from SLAM~\cite{simple-asr-llm}. 
We use the AdamW optimizer~\cite{adamw} with the following hyperparameters: lr = 1.0e-05, beta = (0.9, 0.99), eps = 1.0e-06, and weight\textunderscore{}decay = 0.01. To mitigate potential issues related to gradient explosion during training, we apply gradient clipping~\cite{grad_clip} with a threshold value set to 5. This ensures that gradients over 5 or below -5 are clipped to 5 or -5, respectively. Additionally, we use a gradient accumulation of 20 to increase the effective batch size to 100. 
Regarding the training approach, when training the LLM, we freeze the LLM body and only update the LLM using LoRA fine-tuning~\cite{lora}. We configure LoRA with alpha = 32, rank = 12. The alpha parameter controls the weight of the LoRA matrix, and the rank parameter determines the dimension of the low-rank matrix.  All experiments in this work follow these configurations unless otherwise specified.

\section{Experimental Results}

\begin{table*}[h]
\centering
\setlength{\belowcaptionskip}{0.05cm} 
\setlength{\abovecaptionskip}{0.15cm} 
\caption{Performance CER (\%) on different LLMs through 4-stage fine-tuning on a 40,000-hour multi-accent, multi-dialect dataset.}
\label{librispeech}
\resizebox{1.0\textwidth}{!}{
\begin{tabular}{cccccccccccc}
\hline
\begin{tabular}[c]{@{}c@{}}LLM/\\ Baseline Model\end{tabular} & CTC Finetuned encoder & Projector & LLM finetune method & Frame rate & He Nan         & Shang Hai      & Cantonese      & Hu Nan         & Kespeech      & Test-net      & Test-meeting  \\ \hline
Qwen2-Audio-Instruct                                          & -                     & -         & -                   & 50Hz       & 45.06          & 49.89          & 17.72          & 66.19          & 10.88         & 13.01         & 26.12         \\
Whisper-Large-V3                                              & -                     & -         & -                   & 50Hz       & 87.78          & 87.78          & 55.51          & 83.61          & 31.06         & 8.20          & 9.08          \\
MMGER                                                         & -                     & -         & -                   & -          & -              & -              & -              & -              & 7.52          & -             & -             \\ \hline
Qwen2 0.5b                                                    & True                  & Conv1D      & LoRA                & 12.5Hz     & 20.73          & 21.06          & 11.61          & 26.93          & 7.85          & 7.39          & 8.05          \\
Qwen2 0.5b                                                    & True                  & Conv1D      & LoRA                & 6.25Hz     & 20.48          & 19.81          & 11.03          & 25.16          & 7.64          & 7.29          & 7.69          \\
Qwen2 0.5b                                                    & True                  & Conv1D      & Full                & 6.25Hz     & 19.97          & \textbf{14.58} & 11.73          & 23.89          & 7.20           & 6.80          & 6.98          \\
Qwen2 1.5b                                                    & True                  & Conv1D      & LoRA                & 6.25Hz     & 20.64          & 18.56          & 11.40           & 23.93          & 7.62          & 6.93          & 7.44          \\
Qwen2.5 3b                                                    & False                 & Linear    & LoRA                & 6.25Hz     & 28.06          & 21.09          & 11.06          & 27.54          & 8.27          & 6.91          & 7.23          \\
Qwen2.5 3b                                                    & True                  & Linear    & LoRA                & 6.25Hz     & 27.72          & 21.78          & 11.79          & 25.44          & 7.78          & 7.22          & 7.19          \\
Qwen2.5 3b                                                    & True                  & Conv1D      & LoRA                & 6.25Hz     & 19.95          & 17.33          & 11.31          & 24.31          & 6.84          & \textbf{6.59} & 6.99          \\
Qwen2 7b                                                      & True                  & Conv1D      & LoRA                & 6.25Hz     & \textbf{18.95} & 18.30           & \textbf{10.98} & \textbf{23.31} & \textbf{6.48} & 7.54          & \textbf{6.77} \\ \hline
\end{tabular}

}
\label{tab:40000}
\vspace{-0.1cm} 
\end{table*}

\begin{table}[]
  \centering
\setlength{\belowcaptionskip}{0.05cm} 
\setlength{\abovecaptionskip}{0.15cm} 
\caption{
The improvement in CER (\%) of the model across multiple dialectal datasets in 4-stage fine-tuning.}
\resizebox{0.5\textwidth}{!}{
\begin{tabular}{ccccccccc}
\hline
\multirow{2}{*}{Stage} & \multirow{2}{*}{Projector} & \multirow{2}{*}{He Nan} & \multirow{2}{*}{Shang Hai} & \multirow{2}{*}{Hu Nan} & \multirow{2}{*}{Cantonese} & \multirow{2}{*}{kespeech} & \multirow{2}{*}{Test net} & \multirow{2}{*}{Test meeting} \\
                       &                            &                         &                            &                         &                            &                           &                           &                               \\ \hline
\multirow{2}{*}{1}     & Linear                     & 58.59                   & 84.44                      & 115.68                  & 25.32                      & 25.74                     & 17.68                     & 20.82                         \\
                       & Conv1d                     & 61.13                   & 58.16                      & 169.03                  & 21.35                      & 24.86                     & 21.89                     & 22.41                         \\
\multirow{2}{*}{2}     & Linear                     & 60.23                   & 61.25                      & 72.37                   & 21.89                      & 21.01                     & 14.73                     & 15.60                         \\
                       & Conv1d                     & 56.49                   & 50.22                      & 176.89                  & 17.00                      & 18.38                     & 16.71                     & 15.14                         \\
\multirow{2}{*}{3}     & Linear                     & 32.75                   & 25.44                      & 23.90                   & 12.22                      & 8.90                      & 8.04                      & 8.05                          \\
                       & Conv1d                     & 27.37                   & 25.16                      & 20.45                   & 11.73                      & 8.42                      & 7.48                      & 8.01                          \\
\multirow{2}{*}{4}     & Linear                     & 27.72                   & 21.78                      & 21.78                   & \textbf{11.30}             & 7.90                      & 6.96                      & 7.36                          \\
                       & Conv1d                     & \textbf{25.98}          & \textbf{20.19}             & \textbf{18.56}          & 11.39                      & \textbf{7.40}             & \textbf{6.84}             & \textbf{7.12}                 \\ \hline
\end{tabular}
}
\label{tab:stagesrtest}
\vspace{0cm} 
\end{table}
\subsection{Comparison of Projector Architectures}
We evaluate the effectiveness of different projection layers through small-scale experiments. Specifically, four types of projection layers—Linear, Conv1d, Transformer, and Q-Former are compared using the Data2Vec2 model pre-trained on 300,000 hours of data and tested on the Test-meeting and Test-net datasets.
In these experiments, the LLM is fixed as Qwen2 0.5B, and training is limited only to the first stage for 1,000,000 steps. The Data2vec2 encoder, originally at a sampling rate of 25Hz, is further downsampled by a factor of 4, resulting in an embedding sampling rate of 6.25Hz at the LLM input. For the Q-Former projection layer, a fixed embedding length of 64 is used instead of a predefined sampling rate.
The results, presented in Table \ref{tab:projectorsrtest}, indicate that among the four projection layers, the fully connected layer achieves the best overall performance.

\subsection{ Projector Sampling Rate}
We explore the impact of varying down-sampling rates in the projector.  Experiments were conducted on projection layers implemented using fully connected layers, convolutional layers, and Transformer layers, as shown in table~\ref{tab:projectorsrtest}. For Q-Former, which compresses speech representations into a fixed-length embedding, comparisons with specific sampling rates are not feasible.
Specifically, we test four down-sampling rates—1, 2, 4, and 8—corresponding to frame rates of 25 Hz, 12.5 Hz, 6.25 Hz, and 3.125 Hz for the speech encoder output. In contrast, Q-Former is fixed to produce embeddings with a length of 64. The results reveal that lower down-sampling rates preserve more information, resulting in lower CER. However, lower down-sampling rates also increase the computational load on the LLM, requiring more resources during both training and inference.
Subsequent experiments use a down-sampling rate of 4 to balance recognition accuracy and computational efficiency. Combined with the encoder's intrinsic down-sampling rate of 4, this configuration has a frame rate of 6.25 Hz for the LLM input, offering a practical trade-off between performance and resource consumption.

\subsection{Comparison of ASR-Finetuned and Non-Finetuned Encoders}
We investigate whether ASR-specific finetuning of the encoder, using the CTC objective, benefits the semantic alignment of the encoder with the LLM or introduces conflicts with the knowledge required during LLM finetuning. To this end, we pre-train a Data2Vec2 encoder and subsequently finetune it with CTC to acquire speech recognition capabilities.
Specifically, the Data2Vec2 model is finetuned for 10 epochs using the same supervised 40,000-hour dataset. The resulting ASR-finetuned encoder is then used to replace the pre-trained encoder. Each setup follows the same 1,000,000-step finetuning process.
As shown in Table \ref{tab:projectortest}, the results indicate that ASR-finetuned encoders perform worse across all four projection layer types compared to their non-finetuned counterparts. Notably, the performance gap was most significant in the case of Conv1d projection layers. This suggests that the knowledge obtained through CTC finetuning may conflict with the semantic information required during LLM finetuning, ultimately hindering the overall recognition performance.

\subsection{Analysis of Multi-Stage Finetuning}
We extend the investigation to large-scale experiments with multi-stage training. Specifically, we select the configurations with the best and worst (Linear and Conv1d) performance after the first training stage, as identified in Table \ref{tab:projectortest}, and use them for the 4-stage finetuning process using the larger supervised dataset containing 40,000 hours of multi-dialect speech data. The results are presented in Table \ref{tab:stagesrtest}.
The findings indicate that although the performance gap between the best and worst configurations is substantial after the first stage (31.46/27.87 vs. 16.47/8.80 on the Test-net and Test-meeting datasets), this disparity progressively decreases during the subsequent training stages and is even reversed by the final stage. Based on these results, we hypothesize that as additional model components are unfrozen during multistage fine-tuning, the differences in the projector's parameters and structures become increasingly negligible. 


\subsection{Comparative Analysis of LLMs with Different Sizes}
We evaluate the performance of Qwen series LLMs of 0.5B, 1.5B, 3B, and 7B during the 4-stage fine-tuning on the 40,000-hour dataset. The results are summarized in Table~\ref{tab:40000}.
For Qwen2 0.5B, we compare LoRA fine-tuning versus full fine-tuning in LLM. We observe a significant performance increase, which suggests that the LoRA parameter is not sufficient for acoustic representation training. 
In particular, the 3B model belongs to the Qwen2.5 series, which became available only later in the study. Its performance is compared with that of other Qwen2 models. Although Qwen2.5 3B is trained using more text tokens (7T vs. 18T) and has an MMLU close to Qwen2 7B (65.6 vs. 70.3), we observed that this has no significant benefit.
In all experiments, larger LLMs demonstrated superior overall performance, highlighting the benefits of scaling up the model size. However, even the smallest model, the 0.5B LLM, achieved competitive results, surpassing the current SOTA MMGER~\cite{mu2024mmger} in Kespeech, and outperformed Qwen2-Audio and Whisper-Large-V3 in both internal test sets and public test sets.

\section{Conclusion}
In this study, we investigate the performance of various structural configurations within the paradigm of self-supervised training encoders paired with LLM decoders, leveraging a large-scale dataset comprising 40,000 hours of Chinese dialect and accented speech. For the encoder, we employ Data2Vec2, pre-trained on 300,000 hours of unlabeled dialect and accented speech data. Regarding the projection layer, our finding reveals that fully connected layers demonstrated a performance advantage over other types of projectors. In terms of training strategies, we observe that multi-stage training significantly improved model performance. Extending the conventional 3-stage training approach by incorporating a fourth stage further enhanced the results. The distinction in the effectiveness of the projector is most pronounced during the first stage of training, while these differences diminish progressively as the parameters of the pre-trained model unfreeze throughout the multi-stage training process. We achieve SOTA performance on multiple Chinese dialect test sets including Kespeech, and we will open-source our recipes and pre-trained models.

\bibliographystyle{IEEEtran}
\bibliography{mybib}

\end{document}